\documentclass[letterpaper]{article} % DO NOT CHANGE THIS
\usepackage{aaai2026}  % DO NOT CHANGE THIS
\usepackage{times}  % DO NOT CHANGE THIS
\usepackage{helvet}  % DO NOT CHANGE THIS
\usepackage{courier}  % DO NOT CHANGE THIS
\usepackage[hyphens]{url}  % DO NOT CHANGE THIS
\usepackage{graphicx} % DO NOT CHANGE THIS
\usepackage{natbib}  % DO NOT CHANGE THIS AND DO NOT ADD ANY OPTIONS TO IT
\usepackage{caption} % DO NOT CHANGE THIS AND DO NOT ADD ANY OPTIONS TO IT
\usepackage{amsmath} % Added for split environment
\usepackage{multirow}
\usepackage{booktabs}
\usepackage{amssymb}
\usepackage{algorithm}
\usepackage{algorithmic}
\usepackage{newfloat}
\usepackage{listings}

\DeclareCaptionStyle{ruled}{labelfont=normalfont,labelsep=colon,strut=off} % DO NOT CHANGE THIS
\floatstyle{ruled}
\newfloat{listing}{tb}{lst}{}
\floatname{listing}{Listing}
\nocopyright

\title{Gradient as Conditions: Rethinking HOG for All-in-one Image Restoration}
\author {
    Jiawei Wu\textsuperscript{\rm 1},
    Zhifei Yang\textsuperscript{\rm 2},
    Zhe Wang\textsuperscript{\rm 1},
    Zhi Jin\textsuperscript{\rm 1,3}*,
}
\affiliations {
    \textsuperscript{\rm 1}School of Intelligent Systems Engineering, Shenzhen Campus of Sun Yat-sen University, Guangdong, China.\\
    \textsuperscript{\rm 2}School of Computer Science, Peking University, China.\\
    \textsuperscript{\rm 3}Guangdong Provincial Key Laboratory of Fire Science and Intelligent Emergency Technology, Shenzhen, China.
    wujw97@mail2.sysu.edu.cn
}

\begin{document}

\maketitle

\begin{abstract}
All-in-one image restoration (AIR) aims to address diverse degradations within a unified model by leveraging informative degradation conditions to guide the restoration process. However, existing methods often rely on implicitly learned priors, which may entangle feature representations and hinder performance in complex or unseen scenarios. Histogram of Oriented Gradients (HOG) as a classical gradient representation, we observe that it has strong discriminative capability across diverse degradations, making it a powerful and interpretable prior for AIR. Based on this insight, we propose HOGformer, a Transformer-based model that integrates learnable HOG features for degradation-aware restoration. The core of HOGformer is a Dynamic HOG-aware Self-Attention (DHOGSA) mechanism, which adaptively models long-range spatial dependencies conditioned on degradation-specific cues encoded by HOG descriptors. To further adapt the heterogeneity of degradations in AIR, we propose a Dynamic Interaction Feed-Forward (DIFF) module that facilitates channel–spatial interactions, enabling robust feature transformation under diverse degradations. Besides, we propose a HOG loss to explicitly enhance structural fidelity and edge sharpness. Extensive experiments on a variety of benchmarks, including adverse weather and natural degradations, demonstrate that HOGformer achieves state-of-the-art performance and generalizes well to complex real-world scenarios. Code is available at \url{https://github.com/Fire-friend/HOGformer}.
\end{abstract}

% Uncomment the following to link to your code, datasets, an extended version or similar.
% You must keep this block between (not within) the abstract and the main body of the paper.
% \begin{links}
%     \link{Code}{https://aaai.org/example/code}
%     \link{Datasets}{https://aaai.org/example/datasets}
%     \link{Extended version}{https://aaai.org/example/extended-version}
% \end{links}

\section{Introduction}
\begin{figure*}[t]
    \centering
    \includegraphics[width=\linewidth]{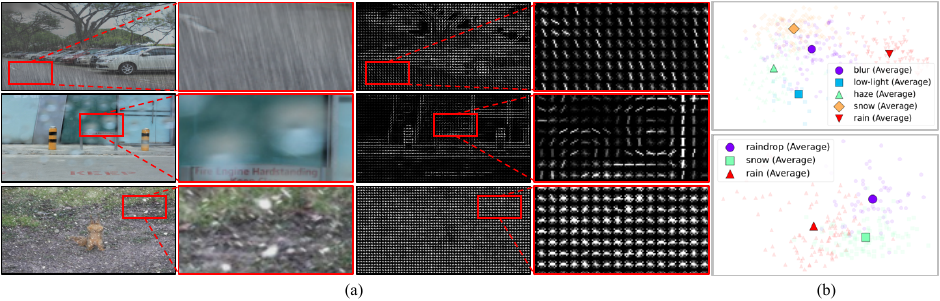}
    \caption{Visualization of HOG feature distributions under various degradations. (a) Example images of different weather conditions with corresponding HOG feature visualizations. (b) HOG features for five natural degradations \cite{zheng2024selective} and three adverse weather degradations \cite{sun2024restoring}, using 100 randomly selected images for each degradation. }
    \label{fig:motivetion}
\end{figure*}
Real-world images are frequently affected by diverse degradations such as blur, low-light, and adverse weather. While task-specific deep learning models have achieved remarkable progress in restoring images under individual degradations \cite{he2024latent,wang2023fourllie,jin2025mb}, they often require separate models, leading to high training and deployment overhead. AIR \cite{jiang2024survey} offers a practical alternative by handling diverse degradations in a single model, which is critical for real-world applications like autonomous driving \cite{chen2024end}.

A key challenge in AIR is designing a model that can adaptively respond to the specific degradation present in each input. To address this, recent works \cite{li2022all,potlapalli2023promptir,ai2024multimodal,jiang2024autodir} have proposed shared backbone architectures modulated by degradation-aware conditional features, enabling a unified framework to handle diverse tasks. Among them, prompt-based methods such as PromptIR \cite{potlapalli2023promptir} inject degradation-specific prompts to guide the network behavior, while others exploit multimodal inputs \cite{ai2024multimodal} to capture complementary degradation cues. Despite their promise, the success of these methods critically depends on the quality and reliability of the conditioning. In practice, such signals are often derived from implicitly learned priors, which operate as black boxes with limited interpretability. This lack of transparency makes it difficult to disentangle the underlying degradations, especially when they exhibit subtle or overlapping characteristics. Consequently, feature representations optimized for one degradation (e.g., denoising) may conflict with those required for another (e.g., dehazing), leading to feature entanglement and reduced generalization performance across tasks \cite{li2020learning}. While alternative methods based on Mixture-of-Experts (MoE) \cite{zamfir2024complexity,ai2024lora} offer greater flexibility through expert routing, they come with substantial computational overhead, making them less suitable for efficient deployment. These limitations underscore the requirement for a conditional mechanism that is not only explicit and discriminative, but also computationally efficient and robust to diverse degradations.

In this work, we revisit classical feature descriptors and identify the HOG \cite{dalal2005histograms} features as a surprisingly effective and interpretable prior for AIR. As shown in Figure \ref{fig:motivetion}, we observe that the gradient patterns captured by HOG form distinct signatures for different degradations. For example, in the case of rain (Figure~\ref{fig:motivetion}(a)), falling rain produces vertical streaks that are reflected as strong vertical gradients, while raindrop regions appear as isolated circular patches with low gradient magnitudes. In contrast, snow degradation introduces widespread high-magnitude gradients with low directional variance, forming dense and uniform textures. This discriminative behavior persists across both natural and adverse weather conditions (Figure~\ref{fig:motivetion}(b)). The effectiveness of HOG stems from its two complementary components, i.e., gradient magnitude and orientation, which explicitly encode local intensity variations and directional structure. These observations raise a central question: \textit{Can the inherent degradation-discriminative property of HOG be leveraged to guide the design of AIR networks, enabling them to explicitly capture degradation-specific gradient patterns for more efficient AIR?}

To this end, we propose HOGformer, an AIR framework that integrates HOG as gradient priors in a principled and degradation-aware manner. Instead of incorporating the static handcrafted HOG feature, we reformulate it as a learnable and differentiable module that provides dynamic and context-aware guidance throughout the network. At the core of HOGformer lies the Dynamic HOG-aware Self-Attention (DHOGSA) mechanism, which integrates HOG-derived cues into the self-attention process. This design enables the network to focus on degradation-specific patterns, such as rain streaks or haze-induced texture loss, by explicitly modulating attention maps based on gradient information. To further enhance the adaptability of feature representations, we introduce a Dynamic Interaction Feed-Forward (DIFF) module that facilitates effective channel–spatial interaction, allowing the model to robustly transform features under diverse and heterogeneous degradations. In addition, we propose a dedicated HOG supervision loss that directly constrains the reconstruction of gradient magnitude and orientation. This explicit guidance promotes sharper edges and better preservation of structural details. This unified design enables effective restoration across diverse degradations, achieving state-of-the-art (SOTA) performance on both adverse weather and natural image AIR tasks. Our main contributions are summarized as follows:
% Specifically, we propose a dynamic HOG-aware self-attention (DHOGSA) mechanism that integrates gradient magnitude and orientation information into the self-attention process. By leveraging HOG-derived cues, DHOGSA dynamically adjusts its focus to facilitate the restoration of critical regions, such as vertically aligned streaks caused by rain or spatially uniform textures introduced by haze. 
% To improve the adaptability of feed-forward module to different degradations, we propose an efficient dynamic interaction feed-forward (DIFF) module. This module facilitates global information exchange by utilizing shuffle and gated mechanisms across spatial and channel dimensions. Besides, we propose a dedicated HOG loss supervises both gradient magnitude and orientation, thereby promoting edge sharpness and structural fidelity. By embedding learnable HOG descriptors into the network, HOGformer integrates gradient priors in a principled and degradation-aware manner. This unified design enables effective restoration across diverse degradation types, achieving state-of-the-art performance on both adverse weather and natural image AIR tasks. The primary contributions of this work are as follows:
\begin{itemize}
    \item We identify that classical HOG features serve as an explicit and highly discriminative prior for distinguishing diverse degradations in AIR, offering a compelling alternative to implicit conditional mechanisms.
    \item We propose HOGformer, an AIR network that embeds learnable HOG cues into core network components for degradation-aware adaptation. HOGformer integrates a dynamic HOG-aware self-attention (DHOGSA) mechanism, an efficient dynamic interaction feed-forward (DIFF) module, and a dedicated HOG supervision loss.
    \item Extensive experiments show that HOGformer achieves state-of-the-art performance across various degradations, including adverse weather and natural conditions.
\end{itemize}

\begin{figure*}[t]
    \centering
    \includegraphics[width=\linewidth]{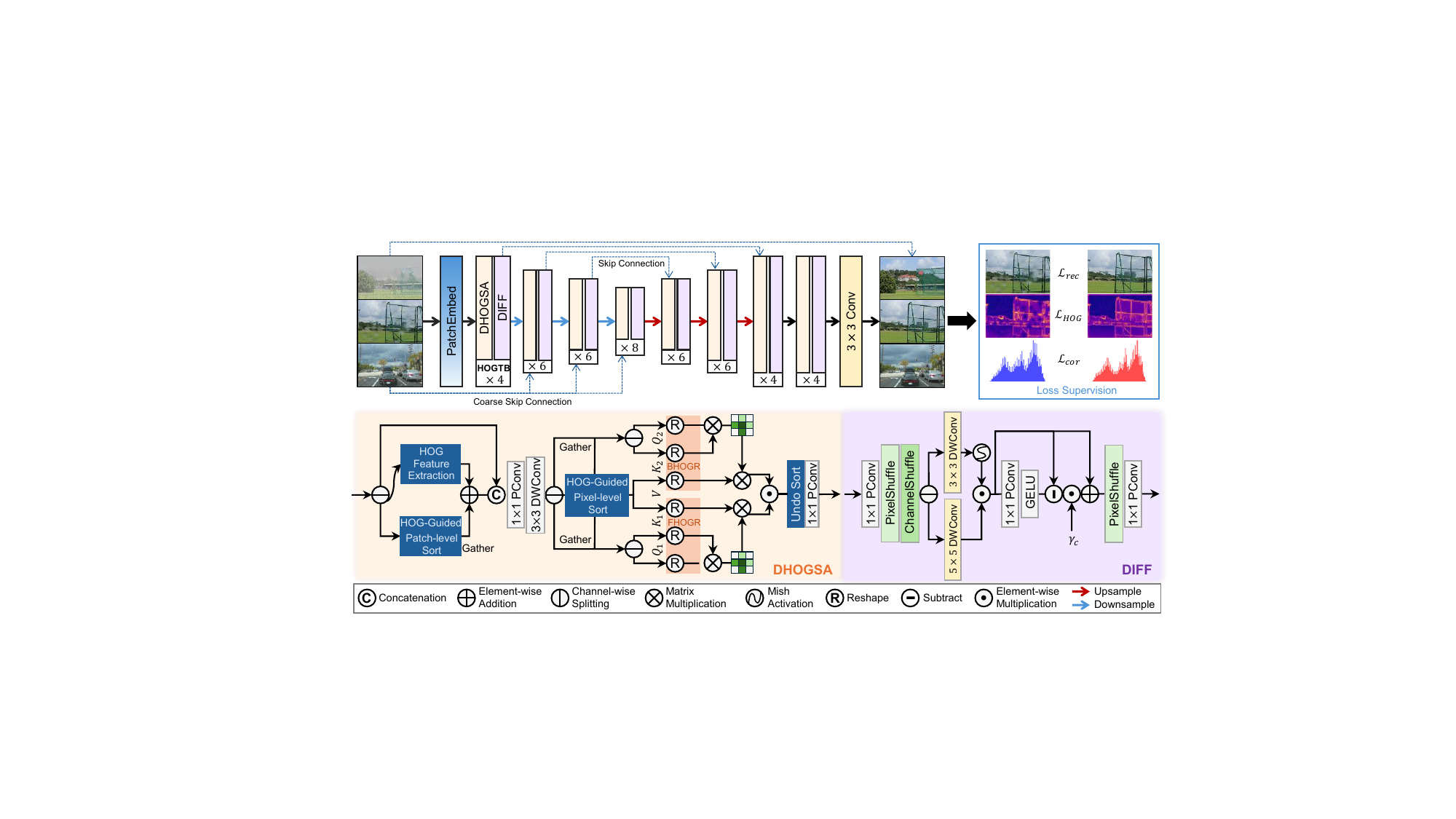}
    \caption{The overall architecture of our HOGformer. It includes the HOG Transformer block with the Dynamic HOG-aware Self-Attention (DHOGSA) module and Dynamic Interaction Feed-Forward (DIFF) module.}
    \label{fig:method}
\end{figure*}
\section{Related Works}
\subsection{Single-Task Image Restoration}
Image restoration is a fundamental problem in computer vision. Traditional methods constrain the solution space using human priors and handcrafted features \cite{banham1997digital}. With deep learning, various approaches have achieved strong performance across tasks \cite{he2024latent,wang2023fourllie,qiu2023mb,zamir2022restormer,guo2024mambair}. Vision Transformers further improve restoration by modeling long-range dependencies \cite{liang2021swinir,zamir2022restormer,wang2022uformer}. For example, Swin-IR \cite{liang2021swinir} adopts a shifted window strategy for local-global context modeling, while Restormer \cite{zamir2022restormer} introduces a multi-stage attention design to balance accuracy and efficiency. Transformer-based models have also been applied to deraining \cite{chen2023learning}, desnowing \cite{chen2022snowformer}, dehazing \cite{song2023vision,zhang2024proxy}, deblurring \cite{liang2024image,kong2023efficient}, and low-light enhancement \cite{cai2023retinexformer,zhang2023lrt}. However, these methods require separate training, increasing computational and deployment costs \cite{sun2024restoring}.

\subsection{All-in-One Image Restoration}
AIR aims to handle diverse degradations using a single model without task-specific retraining \cite{jiang2024survey,li2022all,sun2024restoring,zheng2024selective}. Typically, InstructIR \cite{conde2024instructir} employs natural language instructions to specify restoration goals but incurs high data preparation costs. Methods such as Painter \cite{wang2023images} and DA-CLIP \cite{luo2023controlling} leverage on-the-fly learning to adapt large models, while DiffUIR \cite{zheng2024selective} builds on residual diffusion to address diverse degradations. PromptIR \cite{potlapalli2023promptir} encodes degradation conditions into prompts to guide restoration. These approaches rely on implicitly constructed degradation-specific conditions to enable flexible inference. However, the construction of such conditions is often complex and lacks consistent priors. Recently, Histoformer \cite{sun2024restoring} incorporates grayscale histogram priors into Transformer models and shows strong performance on adverse weather restoration. However, histogram-based features offer limited discriminative power; for example, blurry and clean images can share similar histograms. In this work, we observe that HOG offers a clearer way to distinguish between different degradations. Our method utilizes HOG features to provide explicit guidance that is aware of the degradation throughout the restoration process.

\section{Method}
\subsection{Overall Architecture}
Motivated by the observation (Figure \ref{fig:motivetion}) that the HOG effectively captures degradation-specific patterns, we propose HOGformer, an AIR network that explicitly incorporates HOG-derived cues into the restoration process. As shown in Figure~\ref{fig:method}, HOGformer employs a multi-stage U-shaped encoder-decoder architecture to progressively model and restore diverse degradations within a single model.

The encoder begins by applying a $3 \times 3$ convolution to extract shallow features from the input image $I^{lq}$. These features are successively downsampled and enriched through stacked HOG Transformer Blocks (HOGTBs), which serve as the core feature extractors. To retain low-frequency priors and facilitate residual learning, coarse skip connections~\cite{sun2024restoring} are introduced to highlight degradation residuals, incorporating average pooling, pointwise convolution, and depthwise convolution. Downsampling and upsampling are achieved using pixel-unshuffle and pixel-shuffle, respectively, which ensure efficient resolution transitions while preserving structural consistency. Additionally, symmetric skip connections between the encoder and decoder further enhance information flow and detail recovery. The core of the HOGformer lies the HOGTBs, which incorporate dynamic HOG-aware self-attention (\textbf{DHOGSA}) and dynamic interaction feed-forward (\textbf{DIFF}) as illustrated in Figure~\ref{fig:method}. DHOGSA utilizes gradient information to adjust attention through HOG-guided sorting at both pixel- and patch-level. This facilitates selective focus on regions sensitive to degradation regions. DIFF facilitates effective channel–spatial interaction to enable robust transformation of features under diverse degradations. Each HOGTB employs a residual paradigm to ensure stable optimization as follows:
\begin{equation}
\begin{split}
\mathbf{F}'_l &= \mathbf{F}_{l-1} + \text{DHOGSA}(\text{LN}(\mathbf{F}_{l-1})), \\
\mathbf{F}_l &= \mathbf{F}'_l + \text{DIFF}(\text{LN}(\mathbf{F}'_l)),
\end{split}
\end{equation}
where $F_l$ denotes the output of the $l$-th layer, LN is the layer normalization. This structured integration of HOG-guided attention and interaction enables efficient AIR.

\subsection{Dynamic HOG-aware Self-Attention (DHOGSA)}
The first core module in each HOGTB is DHOGSA, designed to capture long-range and degradation-specific dependencies using gradient-domain priors. Traditional self-attention mechanisms~\cite{zamir2022restormer} typically operate within fixed windows or channel-wise structures, making them suboptimal for modeling the non-uniform and spatially varying patterns present in diverse degradations. DHOGSA overcomes this by leveraging differentiable HOG descriptors to explicitly sort features based on their gradient magnitudes and orientations before computing attention. This sorting operation is conducted at both patch and pixel levels, grouping spatial locations with similar degradation patterns to facilitate more effective attention computation.

\paragraph{Local Dynamic-range Convolution (LDRConv).} 
To facilitate effective long-range modeling, we begin by enhancing local degradation structures through a Local Dynamic-Range Convolution (LDRConv) module. Unlike standard convolutions with fixed receptive fields, LDRConv dynamically reorganizes features based on their gradient distribution. Specifically, LDRConv performs HOG-guided patch-level sorting, followed by modulation using learnable bin-wise HOG priors. This patch-wise operation groups regions with similar gradient properties while preserving global spatial consistency, in contrast to pixel-wise sorting which may disrupt the overall structure. This design ensures that the model retains critical semantic and geometric information, which is essential for image restoration tasks.
% To enable effective global attention, we first need to extract potent local features that are sensitive to degradation structures. Standard convolutions with fixed receptive fields struggle to adapt to the varied structures of degradations that HOG can distinguish. LDRConv addresses this by using HOG-guided sorting and adding learnable bin-wise HOG features from regions with similar gradient characteristics. This pre-organization makes subsequent attention operations more effective while preserving local structure.

\begin{figure}[t]
    \centering
    \includegraphics[width=\linewidth]{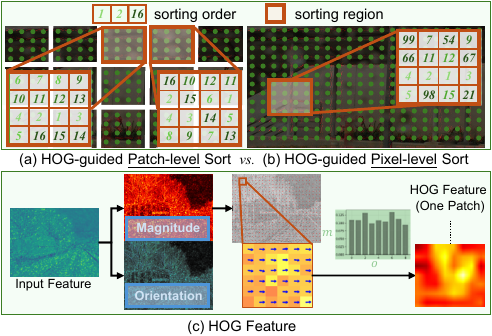}
    \caption{HOG-guided mechanism: (a) Patch-level sorting. (b) Pixel-level sorting. (c) HOG feature extraction process.}
    \label{fig:HOG_sort}
\end{figure}

Given input features $\mathbf{F} \in \mathbb{R}^{C \times H \times W}$, we first compute the gradient magnitude $m$ and orientation $o$ using Sobel filters:
\begin{equation}
    \mathbf{m} = \sqrt{\mathbf{g_x}^2 + \mathbf{g_y}^2},o = \left\lfloor \frac{\operatorname{atan2}(\mathbf{g_y}, \mathbf{g_x}) + \pi}{2\pi} N_{\text{bin}} \right\rfloor,
\end{equation}
where $\mathbf{g_x}$ and $\mathbf{g_y}$ are obtained directional gradients using Sobel operation, $N_{\text{bin}}$ is the number of HOG bins. To preserving the original representation capacity, we split $\mathbf{F}$ into $\mathbf{F}_1, \mathbf{F}_2$ and only apply HOG-guided patch-wise sorting (Figure \ref{fig:HOG_sort}(a)) to $\mathbf{F}_1$, followed by modulation with learnable HOG priors $\operatorname{HOG}_\theta(\mathbf{F}_1)$ to enhance local gradient awareness under diverse degradations:
\begin{equation}
\begin{split}
    &\mathbf{F}_1 = \operatorname{Sort_{patch}}(\mathbf{F}_1) + \operatorname{HOG}_\theta(\mathbf{F}_1), \\
    &\mathbf{F} = \operatorname{Conv_{3\times3}^d}(\operatorname{Conv_{1\times1}^p}(\operatorname{Concat}(\mathbf{F}_1,\mathbf{F}_2))).
\end{split}
\end{equation}
As shown in Figure \ref{fig:HOG_sort}(c), $\operatorname{HOG}_\theta(\mathbf{F}_1)$ is derived by computing gradient-based histograms over local patches, followed by feature projection, more details can be found in the supplementary material. LDRConv enhances degradation sensitivity while maintaining structural coherence, providing a strong basis for the subsequent self-attention mechanism.
\paragraph{HOG-guided Self-Attention.}
With the enhanced representations from LDRConv, we effectively model the crucial long-range dependencies. Specifically, inspired by the observation that distant pixels affected by the same degradation often exhibit similar HOG responses, we sort pixels (Figure \ref{fig:HOG_sort}(b)) based on HOG descriptors ($\mathbf{m} \cdot \mathbf{o}$). 
To spatially align features, we generate sorting indices from the information-rich value features $\mathbf{V}$, and then use these indices to consistently sort the query ($\mathbf{Q}$), key ($\mathbf{K}$), and value ($\mathbf{V}$) features. To capture multi-scale dependencies, we introduce two parallel attention branches with distinct histogram reshaping strategies. Bin-wise Histogram Reshaping (BHOGR) groups sorted pixels into fixed-size bins to extract coarse, large-scale features. Frequency-wise Histogram Reshaping (FHOGR) clusters pixels with similar HOG values to capture fine textures and local variations. BHOGR focuses on capturing large-scale structural artifacts, such as the spatially uniform distribution of haze, whereas FHOGR is tailored to detect fine-scale, repetitive degradations like rain streaks. Fusing both branches enables a comprehensive representation $\mathbf{F}$. The entire process is as follows:
% In order to align the spatial positions of features, we obtain sorting indexes from the value features $\mathbf{V}$ with rich features and consistently sort the query, key, value features $\mathbf{Q},\mathbf{K},\mathbf{V}$ according to this index.
\begin{equation}
\begin{split}
    &\mathbf{idx} = \operatorname{Sort}\Bigl( \mathcal{R}_{C\times H\times W}^{C\times HW}(o (\mathbf{V}) \cdot m(\mathbf{V})) \Bigr),\\
    &\mathbf{Q}_{\{\cdot\}\in \{B, F\}} = \operatorname{Gather}\Bigl( \mathcal{R}_{C\times H\times W}^{C\times HW}(\mathbf{Q}_{\{\cdot\}}),\, \mathbf{idx} \Bigr),\\
    &\mathbf{K}_{\{\cdot\}\in \{B, F\}} = \operatorname{Gather}\Bigl( \mathcal{R}_{C\times H\times W}^{C\times HW}(\mathbf{K}_{\{\cdot\}}),\, \mathbf{idx} \Bigr), \\
    &\mathbf{V} = \operatorname{Gather}\Bigl( \mathcal{R}_{C\times H\times W}^{C\times HW}(\mathbf{V}),\, \mathbf{idx} \Bigr), \\
    &\mathbf{A}_{\{\cdot\}\in \{B, F\}} = \text{Softmax} \left( \frac{\mathcal{R}_{\{\cdot\}}(Q_{\{\cdot\}})\mathcal{R}_{\{\cdot\}}(K_{\{\cdot\}})^{\top}}{\sqrt{\mathbf{K}}} \right), \\
    &\mathbf{F} = \mathbf{A_B} \mathcal{R}_{B}(\mathbf{V}) \odot \mathbf{A_F} \mathcal{R}_{F}(\mathbf{V}).
\end{split}
\end{equation}
where $\mathbf{K}$ is the number of heads, $\mathcal{R}_{i\in \{B,F\}}$ represents the reshaping operation (BHOGR or FHOGR), and $\odot$ is the Hadamard product.
\begin{table*}[t]
    \centering
    % Use 9pt font for better fit
    \small
    % Set a small base column separation; \extracolsep will handle the rest.
    \setlength{\tabcolsep}{3pt}
    % Use tabular* with \textwidth to span both columns.
    % The column spec is preserved from your original table to maintain its unique structure.
    \begin{tabular*}{\textwidth}{@{\extracolsep{\fill}}lcccc|lcc|lcc|cc}
        \hline
        % \rowcolor{gray!20}
        % & \multicolumn{4}{c|}{\textbf{Snow}} & \multicolumn{3}{c|}{\textbf{Rain \& Haze}} & \multicolumn{3}{c|}{ \textbf{Raindrop}} & & \\ 
        % \cline{2-13}
        % \rowcolor{gray!20}
        & \multicolumn{2}{c}{\textbf{Snow100K-S}} & \multicolumn{2}{c|}{\textbf{Snow100K-L}} & &\multicolumn{2}{c|}{\textbf{Outdoor-Rain}} & &\multicolumn{2}{c|}{\textbf{RainDrop}} & \multicolumn{2}{c}{\textbf{Average}}\\
        % \rowcolor{gray!20}
        \textbf{Method}& \textbf{P $\uparrow$} & \textbf{S $\uparrow$} & \textbf{P $\uparrow$} & \textbf{S $\uparrow$} & \textbf{Method}& \textbf{P $\uparrow$} & \textbf{S $\uparrow$} & \textbf{Method}& \textbf{P $\uparrow$} & \textbf{S $\uparrow$} & \textbf{P $\uparrow$} & \textbf{S $\uparrow$}\\
        \hline
        % SPANet & 29.92 & 0.8260 & 23.70 & 0.7930 & CycleGAN & 17.62 & 0.6560 & pix2pix & 28.02 & 0.8547 & -& -\\
        % JSTASR & 31.40 & 0.9012 & 25.32 & 0.8076 & pix2pix & 19.09 & 0.7100 & DuRN & 31.24 & 0.9259 & -& -\\
        RESCAN & 31.51 & 0.9032 & 26.08 & 0.8108 & HRGAN & 21.56 & 0.8550 & RaindropAttn & 31.44 & 0.9263 & -& -\\
        SnowGAN & 32.33 & 0.9500 & 27.17 & 0.8983 & PCNet & 26.19 & 0.9015 & AttentiveGAN & 31.59 & 0.9274 & -& -\\
        DDMSNet & 34.34 & 0.9445 & 28.85 & 0.8772 & MPRNet & 28.03 & 0.9192 & IDT & 31.87 & 0.9313 & -& -\\
        DTANet & 34.79 & 0.9497 & 30.06 & 0.9017 & NAFNet & 29.59 & 0.9027 & MAXIM & 31.87 & 0.9352 & -& -\\
        Restormer & 36.01 & 0.9579 & 30.36 & 0.9068 & Restormer & 30.03 & 0.9215 & Restormer & 32.18 & 0.9408 & -& -\\
        \hline
        All-in-One & - & - & 28.33 & 0.8820 & All-in-One & 24.71 & 0.8980 & All-in-One & 31.12 & 0.9268 & -& -\\
        TransWeather & 32.51 & 0.9341 & 29.31 & 0.8879 & TransWeather & 28.83 & 0.9000 & TransWeather & 30.41 & 0.9157 & 30.27& 0.9094\\
        Chen et al. & 34.42 & 0.9469 & 30.22 & 0.9071 & Chen et al. & 29.27 & 0.9147 & Chen et al. & 31.81 & 0.9309 & 31.43& 0.9249\\
        WGWSNet & 34.31 & 0.9460 & 30.16 & 0.9007 & WGWSNet & 29.32 & 0.9207 & WGWSNet & 32.38 & 0.9378 & 31.54 & 0.9263\\
        WeatherDiff$_{64}$ & 35.83 & 0.9566 & 30.09 & 0.9041 & WeatherDiff$_{64}$ & 29.64 & 0.9312 & WeatherDiff$_{64}$ & 30.71 & 0.9312 & 31.57& 0.9308\\
        WeatherDiff$_{128}$ & 35.02 & 0.9516 & 29.58 & 0.8941 & WeatherDiff$_{128}$ & 29.72 & 0.9216 & WeatherDiff$_{128}$ & 29.66 & 0.9225 & 30.99& 0.9225\\
        AWRCP & 36.92 & 0.9652 & 31.92 & \textbf{0.9344} & AWRCP & 31.39 & 0.9329 & AWRCP & 31.93 & 0.9314 & 33.04& 0.9409\\
        GridFormer & 37.46 & 0.9640 & 31.71 & 0.9231 & GridFormer & 31.87 & 0.9335 & GridFormer & 32.39 & 0.9362 & 33.36 & 0.9392\\
        MoCE-IR & 37.10 & 0.9654 & 31.88 & 0.9234 & MoCE-IR & 31.42 & 0.9334 & MoCE-IR & 32.23 & 0.9386 & 33.16 & 0.9400\\
        Histoformer & \underline{37.41} & \underline{0.9656} & \underline{32.16} & 0.9261 & Histoformer & \underline{32.08} & \underline{0.9389} & Histoformer & \textbf{33.06} & \underline{0.9441} & \underline{33.67} & \underline{0.9436}\\
        \textbf{HOGformer-L} & \textbf{37.93} & \textbf{0.9685} & \textbf{32.41} & \underline{0.9297} & \textbf{HOGformer-L} & \textbf{32.89} & \textbf{0.9460} & \textbf{HOGformer-L} & \underline{32.72} & \textbf{0.9452} & \textbf{33.99} & \textbf{0.9474}\\
        \hline
    \end{tabular*}
    \caption{Quantitative comparisons of adverse weather restoration. The top half of the tables shows results from task-specific methods, while the bottom half displays evaluations of AIR methods. The \textbf{best} and \underline{second-best} results are highlighted.}
    \label{tab:weather_removal}
\end{table*}
\begin{figure*}[t]
    \centering
    \includegraphics[width=\linewidth]{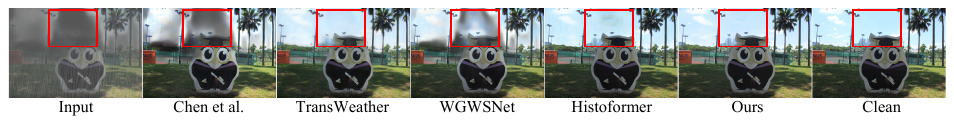}
    \caption{Visual comparison for deraining \cite{sun2024restoring,ozdenizci2023restoring}. Zoom in for the best visualization.}
    \label{fig:com1}
\end{figure*}
\begin{table*}[t]
    \centering
    % Use 9pt font for better fit in a wide table
    \small 
    % Set a base column separation. \extracolsep will manage the rest to fill the width.
    \setlength{\tabcolsep}{4pt}
    % Use tabular* with \textwidth to span both columns and \extracolsep to fill space
    \begin{tabular*}{\textwidth}{@{\extracolsep{\fill}}l|cc|cc|cc|cc|cc|cc}
        \hline
        % \rowcolor{gray!20}
        & \multicolumn{2}{c|}{\textbf{Rain (5 sets)}} & \multicolumn{2}{c|}{\textbf{Low-light}} & \multicolumn{2}{c|}{\textbf{Snow (2 sets)}} & \multicolumn{2}{c|}{\textbf{Haze}} & \multicolumn{2}{c|}{\textbf{Blur}} & \multicolumn{2}{c}{\textbf{Complexity}} \\ 
        % \cline{2-13}
        % \rowcolor{gray!20}
        \textbf{Method}& \textbf{P $\uparrow$} & \textbf{S $\uparrow$} & \textbf{P $\uparrow$} & \textbf{S $\uparrow$} & \textbf{P $\uparrow$} & \textbf{S $\uparrow$} & \textbf{P $\uparrow$} & \textbf{S $\uparrow$} & \textbf{P $\uparrow$} & \textbf{S $\uparrow$} & \textbf{Params (M)} & \textbf{FLOPs (G)}\\ 
        \hline
        SwinIR & 30.78 & 0.923 & 17.81 & 0.723 & - & - & 21.5 & 0.891 & 24.52 & 0.773 & \textbf{0.90} & 752.13\\
        MIRNet-v2 & \underline{33.89} & 0.924 & \textbf{24.74} & 0.851 & - & - & 24.03 & 0.927 & 26.30 & 0.799 & \underline{5.90} & 140.92\\
        % DehazeFormer & - & - & - & - & - & - & \textbf{34.29} & \underline{0.983} & - & - & \underline{4.63} & \underline{48.64}\\
        Restormer & \textbf{33.96} & \textbf{0.935} & 20.41 & 0.806 & - & - & 30.87 & 0.969 & \textbf{32.92} & \textbf{0.961} & 26.12 & 141.00\\
        MAXIM & 33.24 & \underline{0.933} & \underline{23.43} & \underline{0.863} & - & - & \underline{34.19} & \textbf{0.985} & \underline{32.86} & \underline{0.940} & 14.1 & 216.00\\
        % DRSFormer & 33.15 & 0.927 & - & - & - & - & - & - & - & - & 33.7 & 242.9\\
        IR-SDE & - & - & - & - & 20.45 & 0.787 & - & - & 30.70 & 0.901 & 34.20 & \underline{98.30}\\
        WeatherDiff & - & - & - & - & \textbf{33.51} & \textbf{0.939} & - & - & - & - & 82.96 & -\\
        RDDM & 30.74 & 0.903 & 23.22 & \textbf{0.899} & \underline{32.55} & \underline{0.927} & 30.78 & 0.953 & 29.53 & 0.876 & 36.26 & \textbf{9.88}\\ 
        \hline
        % Restormer & 27.10 & 0.843 & 17.63 & 0.542 & 28.61 & 0.876 & 22.79 & 0.706 & 26.36 & 0.814 & 26.12 & 141.00\\
        AirNet & 24.87 & 0.773 & 14.83 & 0.767 & 27.63 & 0.860 & 25.47 & 0.923 & 26.92 & 0.811 & 8.93 & 30.13\\
        Painter & 29.49 & 0.868 & 22.40 & 0.872 & - & - & - & - & - & - & 307.00 & 248.90\\
        % IDR & 29.32 & 0.880 & 21.34 & 0.826 & - & - & 25.24 & 0.943 & 27.87 & 0.846 & 15.34 & -\\
        ProRes & 30.67 & 0.891 & 22.73 & 0.877 & - & - & 32.02 & 0.952 & 27.53 & 0.851 & 307.00 & 248.90\\
        Prompt-IR & 29.56 & 0.888 & 22.89 & 0.847 & - & - & 32.02 & 0.952 & 27.21 & 0.817 & 35.59 & 15.81\\
        DA-CLIP & 28.96 & 0.853 & 24.17 & 0.882 & 30.80 & 0.888 & 31.39 & 0.983 & 25.39 & 0.805 & 174.10 & 118.50\\
        DiffUIR-S & 30.25 & 0.893 & 23.52 & 0.895 & 31.45 & 0.915 &31.83 & 0.954 & 27.79 & 0.830 & \underline{3.27} & \textbf{2.40}\\
        DiffUIR-L & \underline{31.03} & \underline{0.904} & 25.12 & 0.907 & 32.65 & 0.927 &32.94 & 0.956 & \underline{29.17} & \underline{0.864} & 36.26 & \underline{9.88}\\
        \textbf{HOGformer-S}& 30.75& 0.901 & \underline{25.36} & \underline{0.915}& \underline{32.72} & \underline{0.929} & \underline{33.67} & \underline{0.991} & 28.37 & 0.840 & \textbf{2.91} & 20.63\\
        \textbf{HOGformer-L}& \textbf{31.63} & \textbf{0.914} & \textbf{25.57} & \textbf{0.917} & \textbf{34.08} & \textbf{0.941} & \textbf{36.60} & \textbf{0.994} & \textbf{29.95} & \textbf{0.884} & 16.64 & 91.77\\ 
        \hline
    \end{tabular*}%
    \caption{Quantitative results of all-in-one image restoration methods in five tasks. The \textbf{best} and \underline{second-best} results are highlighted. The top is task-specific restoration methods, and the bottom is all-in-one restoration methods.}
    \label{tab:comparison}
\end{table*}
\begin{table*}[t]
    \centering
    \small
    \setlength{\tabcolsep}{4pt}
    \begin{tabular*}{\textwidth}{@{\extracolsep{\fill}}l|cc|cc|cc|cc|cc|cc}
        \hline
        & \multicolumn{2}{c|}{\textbf{Dehazing}} 
        & \multicolumn{2}{c|}{\textbf{Deraining}} 
        & \multicolumn{6}{c|}{\textbf{Denoising on BSD68}} 
        & \multicolumn{2}{c}{\textbf{Average}} \\

        \textbf{Method}
        & \multicolumn{2}{c|}{\textbf{on SOTS}}
        & \multicolumn{2}{c|}{\textbf{on Rain100L}}
        & \multicolumn{2}{c|}{$\boldsymbol{\sigma=15}$}
        & \multicolumn{2}{c|}{$\boldsymbol{\sigma=25}$}
        & \multicolumn{2}{c|}{$\boldsymbol{\sigma=50}$}
        & \multicolumn{2}{c}{ } \\
        
        & \textbf{P $\uparrow$} & \textbf{S $\uparrow$}
        & \textbf{P $\uparrow$} & \textbf{S $\uparrow$}
        & \textbf{P $\uparrow$} & \textbf{S $\uparrow$}
        & \textbf{P $\uparrow$} & \textbf{S $\uparrow$}
        & \textbf{P $\uparrow$} & \textbf{S $\uparrow$}
        & \textbf{P $\uparrow$} & \textbf{S $\uparrow$} \\
        \hline

        BRDNet
        & 23.23 & 0.895 & 27.42 & 0.895 & 32.26 & 0.898 & 29.76 & 0.836 & 26.34 & 0.693 & 27.80 & 0.843 \\
        LPNet
        & 20.84 & 0.828 & 24.88 & 0.784 & 26.47 & 0.778 & 24.77 & 0.748 & 21.26 & 0.552 & 23.64 & 0.738 \\
        FDGAN
        & 24.71 & 0.929 & 29.89 & 0.933 & 30.25 & 0.910 & 28.81 & 0.868 & 26.43 & 0.776 & 28.02 & 0.883 \\
        MPRNet
        & 25.28 & 0.955 & 33.57 & 0.954 & 33.54 & 0.927 & 30.89 & 0.880 & 27.56 & 0.779 & 30.17 & 0.899 \\
        DL
        & 26.92 & 0.931 & 32.62 & 0.931 & 33.05 & 0.914 & 30.41 & 0.861 & 26.90 & 0.740 & 29.98 & 0.876 \\
        AirNet
        & 27.94 & 0.962 & 34.90 & 0.968 & 33.92 & 0.933 & 31.26 & 0.888 & 28.00 & 0.797 & 31.20 & 0.910 \\
        PromptIR
        & 30.58 & 0.974 & 36.37 & 0.972 & 33.98 & 0.933 & 31.31 & 0.888 & 28.06 & 0.799 & 32.06 & 0.913 \\
        AdaIR
        & 31.06 & 0.980 & \textbf{38.64} & 0.983 & \textbf{34.12} & 0.935 & \textbf{31.45} & 0.892 & \textbf{28.19} & 0.802 & 32.69 & 0.918 \\
        \textbf{HOGformer-L}
        & \textbf{31.91} & \textbf{0.981}
        & 38.50 & \textbf{0.983}
        & 34.04 & \textbf{0.935}
        & 31.40 & \textbf{0.892}
        & 28.16 & \textbf{0.804}
        & \textbf{32.80} & \textbf{0.919} \\
        \hline
    \end{tabular*}
    \caption{Comparisons under the three-degradation all-in-one setting.}
    \label{tab:adair_3}
\end{table*}
\begin{table*}[t]
    \centering
    \small
    \setlength{\tabcolsep}{4pt}
    \begin{tabular*}{\textwidth}{@{\extracolsep{\fill}}l|cc|cc|cc|cc|cc|cc}
        \hline
        & \multicolumn{2}{c|}{\textbf{Dehazing}} 
        & \multicolumn{2}{c|}{\textbf{Deraining}} 
        & \multicolumn{2}{c|}{\textbf{Denoising}} 
        & \multicolumn{2}{c|}{\textbf{Deblurring}} 
        & \multicolumn{2}{c|}{\textbf{Low-Light}} 
        & \multicolumn{2}{c}{\textbf{Average}} \\

        \textbf{Method}
        & \multicolumn{2}{c|}{\textbf{on SOTS}}
        & \multicolumn{2}{c|}{\textbf{on Rain100L}}
        & \multicolumn{2}{c|}{\textbf{on BSD68}}
        & \multicolumn{2}{c|}{\textbf{on GoPro}}
        & \multicolumn{2}{c|}{\textbf{on LOL}}
        & \multicolumn{2}{c}{ } \\
        
        & \textbf{P $\uparrow$} & \textbf{S $\uparrow$}
        & \textbf{P $\uparrow$} & \textbf{S $\uparrow$}
        & \textbf{P $\uparrow$} & \textbf{S $\uparrow$}
        & \textbf{P $\uparrow$} & \textbf{S $\uparrow$}
        & \textbf{P $\uparrow$} & \textbf{S $\uparrow$}
        & \textbf{P $\uparrow$} & \textbf{S $\uparrow$} \\
        \hline

        NAFNet
        & 25.23 & 0.939 & 35.56 & 0.967 & 31.02 & 0.883 & 26.53 & 0.808 & 20.49 & 0.809 & 27.76 & 0.881 \\
        HINet
        & 24.74 & 0.937 & 35.67 & 0.969 & 31.00 & 0.881 & 26.12 & 0.788 & 19.47 & 0.800 & 27.40 & 0.875 \\
        MPRNet
        & 24.27 & 0.937 & 38.16 & 0.981 & 31.35 & 0.889 & 26.87 & 0.823 & 20.84 & 0.824 & 28.27 & 0.890 \\
        DGUNet
        & 24.78 & 0.940 & 36.62 & 0.971 & 31.10 & 0.883 & 27.25 & 0.837 & 21.87 & 0.823 & 28.32 & 0.891 \\
        MIRNetV2
        & 24.03 & 0.927 & 33.89 & 0.954 & 30.97 & 0.881 & 26.30 & 0.799 & 21.52 & 0.815 & 27.34 & 0.875 \\
        SwinIR
        & 21.50 & 0.891 & 30.78 & 0.923 & 30.59 & 0.868 & 24.52 & 0.773 & 17.81 & 0.723 & 25.04 & 0.835 \\
        Restormer
        & 24.09 & 0.927 & 34.81 & 0.962 & 31.49 & 0.884 & 27.22 & 0.829 & 20.41 & 0.806 & 27.60 & 0.881 \\
        \hline
        DL
        & 20.54 & 0.826 & 21.96 & 0.762 & 23.09 & 0.745 & 19.86 & 0.672 & 19.83 & 0.712 & 21.05 & 0.743 \\
        TransWeather
        & 21.32 & 0.885 & 29.43 & 0.905 & 29.00 & 0.841 & 25.12 & 0.757 & 21.21 & 0.792 & 25.22 & 0.836 \\
        TAPE
        & 22.16 & 0.861 & 29.67 & 0.904 & 30.18 & 0.855 & 24.47 & 0.763 & 18.97 & 0.621 & 25.09 & 0.801 \\
        AirNet
        & 21.04 & 0.884 & 32.98 & 0.951 & 30.91 & 0.882 & 24.35 & 0.781 & 18.18 & 0.735 & 25.49 & 0.846 \\
        IDR
        & 25.24 & 0.943 & 35.63 & 0.965 & 31.60 & 0.887 & 27.87 & 0.846 & 21.34 & 0.826 & 28.34 & 0.893 \\
        AdaIR
        & 30.53 & 0.978 & 38.02 & 0.981 & 31.35 & 0.889 & 28.12 & 0.858 & 23.00 & 0.845 & 30.20 & 0.910 \\
        \textbf{HOGformer-L}
        & \textbf{31.16} & \textbf{0.979}
        & \textbf{38.05} & \textbf{0.981}
        & \textbf{31.19} & \textbf{0.884}
        & \textbf{28.62} & \textbf{0.867}
        & \textbf{24.46} & \textbf{0.858}
        & \textbf{30.70} & \textbf{0.914} \\
        \hline
    \end{tabular*}
    \caption{Comparisons for five-degradation all-in-one restoration. Denoising results are reported at a noise level of $\sigma = 25$. Methods in the top super-row represent general image restoration methods, while the remaining methods are all-in-one methods.}
    \label{tab:adair_5}
\end{table*}

\subsection{Dynamic Interaction Feed-Forward (DIFF)}
After DHOGSA aggregates spatial features guided by HOG cues, the resulting representations are fed into the proposed Dynamic Interaction Feed-Forward (DIFF) network. Unlike standard, content-agnostic FFNs, DIFF performs dynamic spatial-channel interaction to adaptively refine features based on the content, so that to enhance heavily degraded regions while preserving clean areas. Three complementary designs make it possible: firstly, multi-scale spatial context via parallel branches is captured; secondly, a pixel-wise gating mechanism modulates spatial responses conditioned on input features; thirdly, channel shuffling and aggregation promote cross-channel communication. Therefore, DIFF can achieve content-aware refinement across spatial and channel dimensions. Further details are provided in the supplementary material.

\subsection{Loss Supervision}
To fully exploit the HOG-guided attention and dynamic feature interaction in HOGformer, we design a tailored optimization objective. While pixel-wise losses such as the $L_1$ loss ensure reconstruction accuracy, they often overlook structural and textural fidelity, resulting in overly smooth outputs. To address this limitation, we formulate a composite loss that aligns with the architectural design:
\begin{equation}
\mathcal{L} = \mathcal{L}_{\text{rec}} + \alpha\mathcal{L}_{\text{cor}} + \beta\mathcal{L}_{\text{HOG}},
\end{equation}
where $\alpha$ and $\beta$ are weighting factors. $\mathcal{L}_{\text{rec}}$ denotes the $L_1$ loss for pixel-level accuracy, $\mathcal{L}_{\text{cor}}$ promotes global photometric consistency via Pearson correlation \cite{sun2024restoring}, and $\mathcal{L}_{\text{HOG}}$ enforces structural alignment by minimizing the $L_2$ distance of HOG features between the output and GT. More details are provided in the supplementary material.
\begin{figure}[t]
    \centering
    \includegraphics[width=\linewidth]{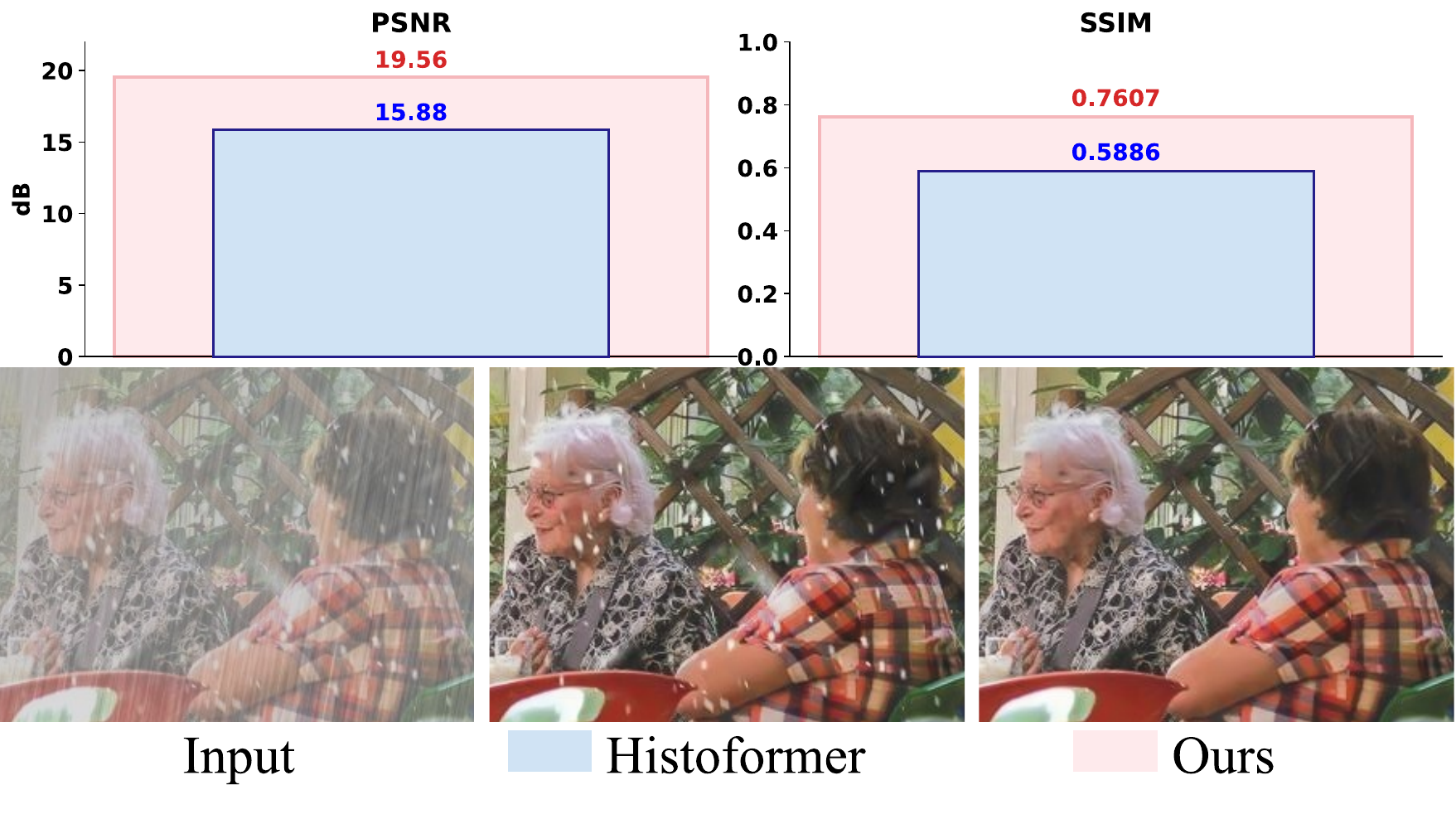}
    % \includesvg[width=\linewidth]{fig/com_hybri.svg}
    \caption{Comparisons on multiple degradations (rain-fog-snow). \textbf{Top:} PSNR and SSIM comparisons between Histformer and our method. \textbf{Bottom:} Visual results on a representative example.}
    \label{fig:com3}
\end{figure}
\section{Experiments}
\subsection{Experimental Setup}
We evaluate HOGformer in four standard AIR settings: (I) 3-task adverse weather removal \cite{sun2024restoring}, including desnowing, draining\&dehazing, and raindrop removal; and (II) 5-task general restoration \cite{zheng2024selective}, including deraining, low-light enhancement, desnowing, dehazing, and deblurring. (III) 3-task general restoration following \cite{potlapalli2023promptir, cui2025adair}. (IV) 5-task general restoration following \cite{potlapalli2023promptir, cui2025adair}.
One AIR model is trained for each setting.

\noindent \textbf{Datasets and Metrics.} 
For Setting (I), we use the AllWeather dataset \cite{sun2024restoring}. For Setting (II), we use task-specific datasets: Merged rain dataset \cite{jiang2020multi, wang2022restoreformer} for deraining, LOL \cite{wei2018deep} for low-light, Snow100K \cite{liu2018desnownet} for desnowing, RESIDE \cite{li2018benchmarking} for dehazing, and GoPro \cite{nah2017deep} for deblurring. To evaluate generalization, we additionally test on real-world datasets: Practical \cite{yang2017deep}, MEF \cite{ma2015perceptual}, NPE \cite{wang2013naturalness}, DICM \cite{lee2013contrast}, HIDE \cite{shen2019human}, RealBlur \cite{rim2020real}, and T-OLED \cite{zhou2021image}. For Settings (III) and (IV), we adopt the same datasets as those used in prior studies \cite{potlapalli2023promptir, cui2025adair}.

For evaluation, we report PSNR \textbf{(P$\uparrow$)} and SSIM \textbf{(S$\uparrow$)} for standard metrics, LPIPS \textbf{(L$\downarrow$)} \cite{zhang2018unreasonable} for perceptual similarity, and NIQE \textbf{(N$\downarrow$)} \cite{mittal2012making} for no-reference assessment.

\subsection{Implementation Details}
We implement our model using PyTorch and conduct experiments on NVIDIA Tesla A100 GPUs. We apply random horizontal and vertical flips for data augmentation. Following Histoformer \cite{sun2024restoring}, $\alpha$ is set to 1, and B equals the number of attention heads. $N_{bin}$ is set to 9 according to original HOG setting \cite{dalal2005histograms}. More details can be seen in the supplementary material.
\begin{table}[t]
    \centering
    % 设置为9pt字号
    \small
    \setlength{\tabcolsep}{4pt}
    % 使用 tabular* 和 \linewidth。总列数从 l+6c 变为 l+5c
    \begin{tabular*}{\linewidth}{@{\extracolsep{\fill}}l|c|c|c|cc}
        \hline
        % \rowcolor{Gray}
         &\textbf{Rain} & \textbf{Low-light} & \textbf{Snow} & \multicolumn{2}{c}{\textbf{Blur}} \\ 
        % \rowcolor{Gray}
         \textbf{Method}& \textbf{N $\downarrow$} & \textbf{N $\downarrow$} & \textbf{N $\downarrow$} & \textbf{P $\uparrow$} & \textbf{S $\uparrow$} \\
        \hline
        WeatherDiff & - & - & \underline{2.96} & - & - \\
        CLIP-LIT & - & \underline{3.70} & - & - & - \\
        Restormer & \underline{3.50} & 3.80 & - & \textbf{32.12} & \textbf{0.926} \\
        RDDM & \textbf{3.34} & \textbf{3.57} & \textbf{2.76} & \underline{30.74} & \underline{0.894} \\
        \hline
        AirNet & 3.55 & 3.45 & 2.75 & 16.78 & 0.628 \\
        Prompt-IR & 3.52 & 3.31 & 2.79 & 22.48 & 0.770 \\
        DA-CLIP & 3.42 & 3.56 & \underline{2.72} & 17.51 & 0.667 \\
        DiffUIR-L & \underline{3.38} & \underline{3.14} & 2.74 & \underline{30.63} & \underline{0.890} \\
        \textbf{HOGformer-L} & \textbf{3.31} & \textbf{3.08} & \textbf{2.69} & \textbf{30.92} & \textbf{0.907} \\
        \hline
    \end{tabular*}
    \caption{Quantitative results of known task generalization. The \textbf{best} and \underline{second-best} results are highlighted. The top is task-specific methods, and the bottom is AIR methods.}
    \label{tab:known}
\end{table}
\begin{table}[t]
    \centering
    % Set font to 9pt for consistency
    \small
    % Set a base column separation; \extracolsep will handle the rest of the spacing.
    \setlength{\tabcolsep}{8pt}
    % Use tabular* to fill the single column width (\linewidth) by stretching inter-column space.
    \begin{tabular*}{\linewidth}{@{\extracolsep{\fill}}l|ccc}
        \hline
        % \rowcolor{gray!20}
         % & \multicolumn{3}{c|}{\textbf{T-OLED}} \\
        % \rowcolor{gray!20}
        \textbf{Method}& \textbf{P $\uparrow$} & \textbf{S $\uparrow$} & \textbf{L $\downarrow$} \\
        \hline
        NAFNet       & \textbf{26.89} & \underline{0.774} & \textbf{0.346} \\
        MPRNet       & 23.33 & \textbf{0.807} & 0.383 \\
        % DGUNet       & 19.67 & 0.627 & 0.570 \\
        % MIRNetV2     & 20.15 & 0.703 & 0.474 \\
        SwinIR       & 17.72 & 0.661 & 0.519 \\
        Restormer    & \underline{20.98} & 0.632 & \underline{0.360} \\
        RDDM         & 17.00 & 0.626 & 0.545 \\
        \hline
        % DL           & 21.23 & 0.656 & 0.434 \\
        % Transweather & 20.52 & 0.666 & 0.451 \\
        AirNet       & 22.73 & 0.739 & 0.374 \\
        IDR          & 27.91 & 0.793 & 0.346 \\
        Prompt-IR    & 20.47 & 0.669 & 0.462 \\
        DA-CLIP      & 15.74 & 0.606 & 0.472 \\
        DiffUIR-L    & \textbf{29.55} & \underline{0.887} & \underline{0.281} \\
        \textbf{HOGformer-L} & \underline{29.33} & \textbf{0.889} & \textbf{0.271} \\
        \hline
    \end{tabular*}
    \caption{Quantitative results of unknown task on T-OLED. The \textbf{best} and \underline{second-best} results are highlighted. The top is task-specific methods, and the bottom is AIR methods.}
    \label{tab:unknown}
\end{table}

% ### **Extracted References**
%   * `\cite{zhou2021image}`
%   * `\cite{chen2022simple}`
%   * `\cite{zamir2021multi}`
%   * `\cite{mou2022deep}`
%   * `\cite{zamir2022learning}`
%   * `\cite{liang2021swinir}`
%   * `\cite{liu2024residual}`
%   * `\cite{zamir2022restormer}`
%   * `\cite{fan2019general}`
%   * `\cite{valanarasu2022transweather}`
%   * `\cite{li2022all}`
%   * `\cite{zhang2023ingredient}`
%   * `\cite{potlapalli2023promptir}`
%   * `\cite{luo2023controlling}`
%   * `\cite{zheng2024selective}`

\subsection{Comparison with State-of-the-Arts}
\noindent \textbf{Setting (I).}
Table~\ref{tab:weather_removal} compares HOGformer with task-specific \cite{li2018recurrent,yang2017deep,zhang2021deep,chen2022simple,zamir2022restormer,li2019heavy,li2016rain,zamir2021multi,chen2022snowformer,qian2018attentive,quan2021removing,tu2022maxim} and all-in-one SOTA methods \cite{li2020all,valanarasu2022transweather,chen2022learning,zhu2023learning,ozdenizci2023restoring,ye2023adverse,wang2024gridformer,sun2024restoring} on synthetic and real-world adverse weather datasets. For fairness, we retrain all-in-one baselines on AllWeather~\cite{li2020all, ozdenizci2023restoring}. HOGformer consistently outperforms existing methods across all degradations. As shown in Figure \ref{fig:com1} and Figure \ref{fig:com3}, HOGformer also restores clearer details and better handles multiple degradations. This underscores the importance of natural discriminative HOG attributes to model degradations.

\noindent \textbf{Setting (II).}
Table~\ref{tab:comparison} compares AIR methods under five natural degradations. HOGformer-S achieves competitive results in low-light enhancement, desnowing, and dehazing, outperforming larger models like DA-CLIP (174.1M). Besides, HOGformer-L achieves best results with reasonable parameters than counterparts \cite{liang2021swinir,zamir2022learning,zamir2022restormer,tu2022maxim,luo2023image,ozdenizci2023restoring,liu2024residual,zamir2022restormer,li2022all,wang2023images,ma2023prores,potlapalli2023promptir,luo2023controlling,zheng2024selective}. These results highlight its balance between performance and complexity.

\noindent \textbf{Setting (III).}
Table~\ref{tab:adair_3} presents quantitative comparisons under the three-degradation all-in-one setting, including dehazing, deraining, and denoising with different noise levels. HOGformer achieves the best average PSNR and SSIM among all methods, and consistently delivers top or near-top performance across all degradation types and noise levels. Compared with recent all-in-one restoration methods such as PromptIR and AdaIR, HOGformer demonstrates superior overall restoration capability and robustness, highlighting its effectiveness in handling multiple degradations within a unified framework.

\noindent \textbf{Setting (IV).}
Table~\ref{tab:adair_5} reports quantitative comparisons of general image restoration methods and all-in-one methods across five degradations, including dehazing, deraining, denoising, deblurring, and low-light enhancement. Among all-in-one approaches, HOGformer consistently achieves the best or highly competitive PSNR and SSIM on all datasets, and obtains the highest average performance overall. Compared with both general restoration models and recent AIR methods, HOGformer demonstrates superior restoration quality across diverse degradations, highlighting its strong generalization ability and balanced effectiveness.

\noindent \textbf{Generalization.} To further assess generalizability, we evaluate HOGformer on both seen and unseen degradations. As shown in Table~\ref{tab:known}, HOGformer consistently outperforms previous state-of-the-art methods on seen tasks.  In Table~\ref{tab:unknown}, HOGformer delivers comparable results on T-OLED with unseen degradations, confirming its robustness and adaptability beyond trained scenarios. This generalization likely stems from the intrinsic properties of HOG features. Unlike learned priors prone to overfitting, HOG captures fundamental structural cues common to many degradations. Therefore, the HOG-guided mechanism can highlight such patterns even in real-world degradations.
\begin{table}[t]
    \centering
    \small
    \setlength{\tabcolsep}{1pt}
    \begin{tabular*}{\linewidth}{@{\extracolsep{\fill}}cccc|ccc}
        \hline
        % \rowcolor{gray!20}
        \textbf{LDRConv} & \textbf{DHOGSA} & \textbf{DIFF} & \textbf{$\mathcal L_{HOG}$} & \textbf{P} $\uparrow$ & \textbf{S} $\uparrow$ & \textbf{Params $\downarrow$} \\
        \hline
        \(\times\) & \(\times\) & \(\times\) & \(\times\) &  30.14&  0.9258 & 14.65M \\
        \checkmark & \(\times\) & \(\times\) & \(\times\) &  30.49&  0.9261 & 14.67M \\
        \checkmark & \checkmark & \(\times\) & \(\times\) &  31.68&  0.9358 & 16.77M \\
        \checkmark & \checkmark & \checkmark & \(\times\) &  32.40&  0.9421 & 16.64M \\
        \checkmark & \checkmark & \checkmark & \checkmark &  \textbf{32.89}&  \textbf{0.9460} & 16.64M \\
        \hline
    \end{tabular*}
    \caption{Ablation study on the proposed core components.}
    \label{tab:ablation}
\end{table}
% \begin{table}[htb]
%     \centering
%     \caption{Ablation study on the FHOGR and BHOGR.}
%     \label{tab:ablation_HOGR}
%     \resizebox{\linewidth}{!}{
%         \setlength{\tabcolsep}{30pt} % Adjust column separation
%         \begin{tabular}{|l|c|c|}
%         \hline
%         \rowcolor{gray!20}
%         \textbf{Method} & \textbf{PSNR (dB)} & \textbf{SSIM} \\
%         \hline
%         w/o FHOGR & 32.59 & 0.9432 \\
%         w/o BHOGR & 31.83 & 0.9386 \\
%         DHOGSA & \textbf{32.89} & \textbf{0.9460} \\
%         \hline
%         \end{tabular}
%     }
% \end{table}
\begin{table}[t]
    \centering
    \small
    \setlength{\tabcolsep}{10pt}
    \begin{tabular*}{\linewidth}{@{\extracolsep{\fill}}l|ccc}
        \hline
        \textbf{} & \textbf{w/o FHOGR} & \textbf{w/o BHOGR} & \textbf{DHOGSA} \\
        \hline
        \textbf{P} $\uparrow$ & 32.59 & 31.83 & \textbf{32.89} \\
        \textbf{S} $\uparrow$ & 0.9432 & 0.9386 & \textbf{0.9460} \\
        \hline
    \end{tabular*}
    \caption{Ablation study on the FHOGR and BHOGR.}
    \label{tab:ablation_HOGR}
\end{table}

\begin{figure}[b]
    \centering
    \includegraphics[width=\linewidth]{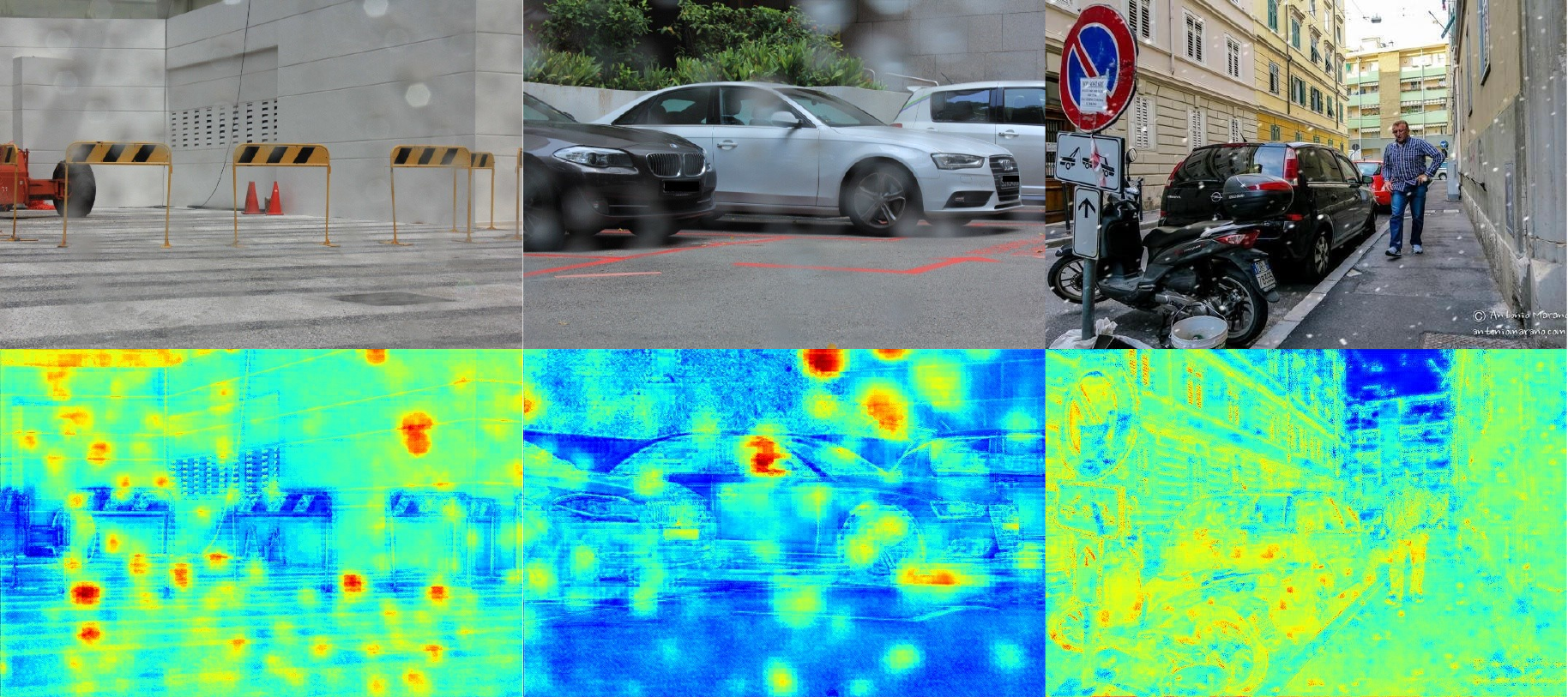}
    % \includesvg[width=\linewidth]{fig/attention}
    \caption{Visualization of the DHOGSA output from the second-to-last HOGTB, highlighting its ability to accurately activate degraded regions.}
    \label{fig:aba_attention}
\end{figure}
\subsection{Ablation Studies}
Ablation studies on Outdoor-Rain are conducted to evaluate our methods and to determine optimal hyperparameters.

\noindent \textbf{Effectiveness of Core Components.} As shown in Table~\ref{tab:ablation}, adding LDRConv significantly improves PSNR and SSIM. This performance gain is attributable to it models local dynamic-range variations commonly found in complex degradations. Introducing DHOGSA further boosts performance by embedding HOG-based priors into the attention mechanism. The reason is that HOG features capture gradient cues that vary across degradations, allowing the model to adjust attention weights adaptively based on perceived degradation as shown in Figure \ref{fig:aba_attention}. The DIFF module brings additional gains by enabling dynamic feature transformation. The reason is that it enhances the ability of model to process regions with varying degradation levels through spatial-channel interaction. Table \ref{tab:ablation_HOGR} shows that both BHOGR and FHOGR are beneficial. We attribute this improvement to BHOGR captures inter-interval dependencies while FHOGR focuses on intra-interval structures. Combining all components yields the best performance, confirming their complementary roles in AIR.

\noindent \textbf{Hyperparameters of HOG Bins.}
Table \ref{tab:hog_bins_ablation} analyzes the effect of the number of bins in extracted HOG features. A small number of bins leads to poor performance because it fails to capture essential gradient orientations necessary for identifying degradations. Increasing the number of bins beyond a certain threshold offers only minor improvements while imposing greater computational overhead. To balance accuracy and efficiency, we set the number of bins to 9. 

\begin{figure}[t]
    \centering
    \includegraphics[width=\linewidth]{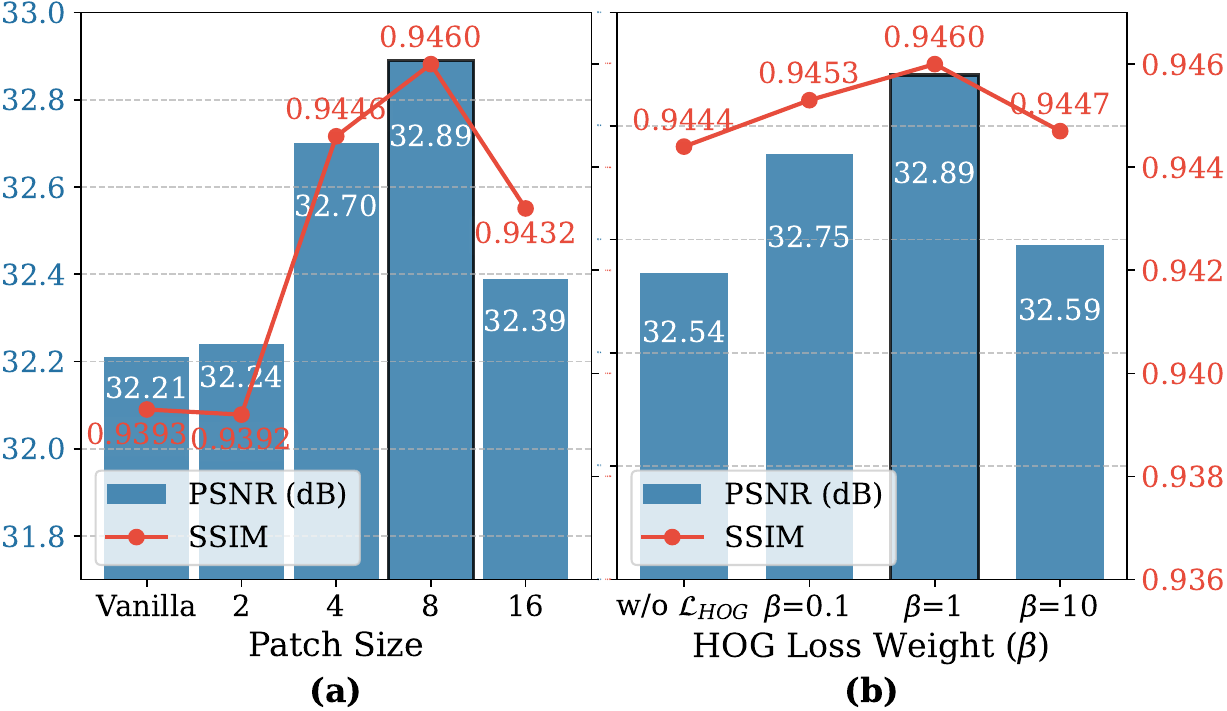}
    \caption{Ablation studies on the patch size of LDRConv (a) and the weight of HOG loss (b).}
    \label{fig:aba_hyper}
\end{figure}

\noindent \textbf{Hyperparameters of Local Dynamic-range Convolution.}
Figure \ref{fig:aba_hyper} (a) shows that LDRConv with patch size 8 achieves the best performance (32.89dB PSNR, 0.9460 SSIM), outperforming the baseline by +0.68dB and +0.0067 SSIM. Smaller patches offer limited context, while overly large patches (e.g., size 16) degrade structural fidelity.

\noindent \textbf{Hyperparameters of HOG Loss.}
As illustrated in Figure \ref{fig:aba_hyper} (b), setting $\beta = 1$ yields the highest performance (32.89dB PSNR, 0.9460 SSIM). A small weight ($\beta = 0.1$) offers marginal gains, while an excessive weight ($\beta = 10$) hampers performance, indicating the need to balance gradient emphasis and overall fidelity.
\begin{table}[t]
    \centering
    % Use 9pt font for consistency with other tables
    \small
    % Set a base column separation; \extracolsep will handle the rest.
    \setlength{\tabcolsep}{6pt} 
    % Use tabular* to fill the single column width (\linewidth)
    \begin{tabular*}{\linewidth}{@{\extracolsep{\fill}}l|cccc}
        \hline
        % \rowcolor{gray!20}
         & \textbf{13 Bins} & \textbf{9 Bins} & \textbf{5 Bins} & \textbf{1 Bin} \\
        \hline
        \textbf{P} $\uparrow$       & 32.93            & 32.89            & 32.64            & 32.32            \\
        % \hline
        \textbf{S} $\uparrow$            & 0.9464           & 0.9460           & 0.9433           & 0.9417           \\
        \hline
    \end{tabular*}
    \caption{Ablation study on the number of bins in extracted HOG features.}
    \label{tab:hog_bins_ablation}
\end{table}
\section{Conclusion}
In this paper, we present HOGformer, an AIR model that leverages the power of HOG features. Through a Dynamic HOG-aware Self-Attention (DHOGSA) mechanism and innovative network components, HOGformer effectively handles diverse image degradations. Our approach demonstrates superior performance while maintaining computational efficiency. Besides, extracting HOG features explicitly helps the model identify various degradations from a gradient perspective and eliminate interfering factors like lighting. This further enhance the generalizability of our method. Future research directions include exploring HOG-based mechanisms in conjunction with emerging architectures (e.g., Mamba) to address the AIR task.

% Additionally, although the proposed framework advances all-in-one degradation handling, there is still room for enhancing the generalization to unseen or more complex degradation scenarios. Generalization to novel degradation patterns persists as a fundamental challenge in all-in-one restoration frameworks. Potential solutions may involve developing robust training paradigms, including adversarial learning and meta-learning strategies, to enhance adaptation to unforeseen degradations.

\bibliography{aaai2026}

\end{document}